\documentclass{Interspeech}
\newcommand{\anonemail}{}

\interspeechcameraready 

\title{NonverbalTTS: A Public English Corpus of Text-Aligned Nonverbal Vocalizations with Emotion Annotations for Text-to-Speech}

\author[affiliation={1}]{Maksim}{Borisov}
\author[affiliation={1}]{Egor }{Spirin}
\author[affiliation={2}]{Daria}{Diatlova}

\affiliation{}{VK Lab}{Russia}
\affiliation{}{Yandex}{Russia}
\keywords{speech synthesis, non-verbal vocalization}

\usepackage{comment}
\usepackage{graphicx} 
\usepackage{xcolor} 
\usepackage{algpseudocode}
\usepackage{algorithm}
\usepackage{tablefootnote}
\DeclareGraphicsExtensions{.png,.pdf,.jpg}

\algnewcommand{\algorithmicforeach}{\textbf{for each}}
\algdef{SE}[FOR]{ForEach}{EndForEach}[1]
  {\algorithmicforeach\ #1\ \algorithmicdo}
  {\algorithmicend\ \algorithmicforeach}

\begin{document}

\maketitle

\begin{abstract}
Current expressive speech synthesis models are constrained by the limited availability of open-source datasets containing diverse nonverbal vocalizations (NVs).
In this work, we introduce NonverbalTTS (NVTTS), a 17-hour open-access dataset annotated with 10 types of NVs (e.g., laughter, coughs) and 8 emotional categories.
The dataset is derived from popular sources, VoxCeleb and Expresso, using automated detection followed by human validation.
We propose a comprehensive pipeline that integrates automatic speech recognition (ASR), NV tagging, emotion classification, and a fusion algorithm to merge transcriptions from multiple annotators.
Fine-tuning open-source text-to-speech (TTS) models on the NVTTS dataset achieves parity with closed-source systems such as CosyVoice2, as measured by both human evaluation and automatic metrics, including speaker similarity and NV fidelity.
By releasing NVTTS and its accompanying annotation guidelines, we address a key bottleneck in expressive TTS research.
The dataset is available at \url{https://huggingface.co/datasets/deepvk/NonverbalTTS}.
    
    
    
\end{abstract}

\section{Introduction}

Expressivity has become a critical component of naturalness in modern text-to-speech (TTS) systems, driving growing interest in emotional and prosody-aware speech synthesis. In addition to fine-grained prosodic variation (e.g., pitch, rhythm, intensity), nonverbal vocalizations (NVs) such as laughter, sighs, and coughs play a crucial role in human communication~\cite{hall2009psychosocial, scherer2011assessing}, yet remain underexplored in controllable speech generation.

Recent expressive TTS systems have begun to incorporate NVs alongside traditional prosodic control. For instance, CosyVoice \cite{du2024cosyvoice} support the generation of NVs like sighs and laughter, FireRedTTS~\cite{guo2024fireredtts} supports the prosodic manipulations such as pitch shifts, syllable elongation and word repetition. EmoCtrl-TTS \cite{wu2024laugh} further extends this to emotionally charged NVs, including crying and moaning.

Despite these advancements, a major limiting factor remains the scarcity of open-source datasets containing diverse and high-quality NVs. Most existing corpora include only basic NVs (e.g., breath, laugh) and frequently suffer from acoustic artifacts, leading many researchers to rely on proprietary datasets with richer annotations \cite{du2024cosyvoice_2, guo2024fireredtts}. The absence of open-source datasets with varied NVs makes reproducing and benchmarking expressive speech synthesis models that can generate NVs extremely difficult.

In this paper, we (1) present the largest open-source dataset for text-to-speech synthesis containing a comprehensive collection of nonverbal vocalizations. We enriched VoxCeleb \cite{nagrani2020voxceleb} and Expresso \cite{nguyen2023expresso} transcriptions by annotating existing NVs, resulting in 10 distinct NV types through automated labeling followed by careful human validation. We (2) provide detailed guidelines for constructing NV-enriched datasets via automated detection and human validation, ensuring reproducibility and scalability. Finally, we (3) conduct an ablation study of the curated dataset, including training a nonverbal synthesis model that achieves performance on par with the current state-of-the-art approach.

\begin{table*}
    \centering

    
    \begin{tabular}{lcccccccc}
\toprule
 & \multicolumn{2}{c}{Num audio} & \multicolumn{2}{c}{Num speakers} & \multicolumn{2}{c}{Duration (h)} & \multicolumn{2}{c}{Num males} \\
 \midrule
Data name & Expresso & VoxCeleb & Expresso & VoxCeleb & Expresso & VoxCeleb & Expresso & VoxCeleb \\
\midrule

Original & 11 954 & 1 281 762 & 4 & 7 365 & 41 & 2 794 & 2 & 4 453 \\

Filtered & 1 804 & 4 452 & 4 & 2 292 & 3.9 & 13.7 & 2 & 1 378 \\

\bottomrule
\end{tabular}

    \caption{Statistics of Expresso and VoxCeleb datasets before filtration (Original) and after filtration. Note that both parts of VoxCeleb are combined.}
    \label{tab:stats}
\end{table*}

\begin{table*}
    \centering
\begin{tabular}{lcccccccccc}
\toprule
NV & Breath & Laugh & Sniff & Cough & Throat & Sigh & Groan & Sneeze & Snore & Grunt \\
\midrule
Count & 3 612 & 1 018 & 388 & 233 & 203 & 145 & 117 & 16 & 13 & 7 \\
\bottomrule
\end{tabular}
\caption{NVs statistics over NVTTS. We can see that Breath and Laugh are the most frequent categories, as expected. However, the overall distribution remains relatively balanced across other NV types. }
\label{tab:nvtts_nv}
\end{table*}

\section{Related work} 
\subsection{Datasets for NV-TTS}\label{sec:related-datasets}
Datasets containing NVs can be categorized into three categories.

In the first category, the AMI Meeting Corpus~\cite{kraaij2005ami} offers 100 hours of meeting recordings with word-level transcriptions, including annotations for laughter and coughs from mostly non-native English speakers. Switchboard~\cite{godfrey1992switchboard} contains 2,400 telephone conversations with annotations for non-verbal elements like laughter and throat-clearing. However, its 8 kHz sampling rate and 1990s telephone quality limit its usefulness for speech synthesis.
Fisher~\cite{cieri2004fisher} includes 16,000 telephone conversations with annotations for sighs, throat clearing, and laughter, but with similar audio quality limitations.

In the second category, more recent TTS-oriented datasets include JNV~\cite{xin2024jvnv}, a Japanese emotional speech corpus with NVs generated via ChatGPT prompting. Despite high recording quality, it has limited speakers.
~\cite{arimoto2012naturalistic} collected emotional speech with non-verbal expressions but only annotated laughter. Expresso\cite{nguyen2023expressobenchmarkanalysisdiscrete} corpus consists of 47 hours of expressive speech from four North American English speakers, recorded in a professional studio with minimal background noise at 48kHz/24bit. Laughter and breathing NVs are annotated and included into transcriptions.

In the third category, large-scale, multilingual, and diverse datasets like Emilia~\cite{he2024emiliaextensivemultilingualdiverse}, which consist of over 101k hours of in-the-wild spontaneous speech rich in non-verbal characteristics, are available. However, these non-verbal elements are not reflected in the textual annotations, making it difficult to achieve high-quality controllable speech synthesis using such data.

In conclusion, existing TTS datasets with non-verbal vocalizations (NVs) either suffer from poor recording quality or lack sufficient annotated data, making them unsuitable for properly training robust, controllable, NV-capable TTS systems.

\subsection{Models for NV-TTS}
Recent TTS systems have demonstrated various approaches to generating NVs.

NVS-TTS \cite{zhang2023nsv} uses self-supervised learning to extract linguistic units for emotion-based TTS with NVs, enabling zero-shot transfer but lacking placement control and relying on private Mandarin data. EmoCtrl-TTS \cite{wu2024laugh} creates emotional speech with NVs for any speaker by conditioning a flow-matching TTS with arousal/valence values and laughter embeddings. Drawing from AMI \cite{kraaij2005ami}, Switchboard \cite{godfrey1992switchboard}, and Fisher \cite{cieri2004fisher} datasets, it offers precise control over laughter position and intensity. FireRedTTS \cite{guo2024fireredtts} converts speech to discrete semantic tokens using a language model approach. Despite impressive results with emotional NVs, it relies on a proprietary 50-hour dataset with minimal documentation and remains closed-source.

CosyVoice \cite{du2024cosyvoice} combines supervised discrete speech tokens with progressive semantic decoding to deliver multilingual synthesis with breathing and laughing capabilities, but uses private NV training data. CosyVoice2 \cite{du2024cosyvoice_2}, an improved streaming version, adds coughs and sighs to its repertoire. While its architecture is open-sourced, the instruction dataset remains proprietary. Our work with NVs builds on this framework.


\section{Dataset Annotation Pipeline}\label{sec:dataset_annotation}

In this section, we describe the general data processing pipeline which we applied to the VoxCeleb \cite{nagrani2020voxceleb} and Expresso \cite{nguyen2023expresso} corpora to construct the NonverbalTTS corpus. The proposed pipeline consists of four main steps: (1) detecting nonverbal vocalizations, (2) identifying emotions, (3) human annotation refinement, and (4) annotation fusion process. 

Since the VoxCeleb dataset \cite{nagrani2020voxceleb} lacks transcriptions, we generated them using the Canary model \cite{Harper_NeMo_a_toolkit}, which currently ranks as the top-performing ASR model according to the Open ASR Leaderboard \cite{open-asr-leaderboard}. Below, we detail each processing step applied to both corpora.

\subsection{Nonverbal Vocalizations Detection}
We employed the BEATs model \cite{schmid2024effective} to detect non-verbal vocalizations in audio samples. We focused on 10 types of NVs: breathing, laughter, sighing, sneezing, coughing, throat clearing, groaning, grunting, snoring and sniffing. These were chosen based on their presence in the AudioSet Ontology \cite{gemmeke2017audio} and how often they appeared in our data.

Since we plan to use human validation in the future, it is acceptable to set the detection threshold at the lowest recommended level of 0.1, as per \cite{schmid2024effective}. With this setting, grunting turned out to be the rarest NV, appearing in only 7 samples.

For creating transcriptions enriched with precise non-verbal vocalization tags, we used the Montreal Forced Aligner (MFA) \cite{mcauliffe2017montreal} to align the audio with its transcription. Leveraging both the alignment and timestamps provided by the event detection model, we accurately placed the non-verbal vocalizations within the transcription. 

\subsection{Emotion Detection}\label{sec:emotion_detection}
We classified all audio samples into 8 emotion categories: angry, disgusted, fearful, happy, neutral, sad, surprised, and other. Although the Expresso dataset \cite{nguyen2023expresso} includes style tags that overlap with emotional categories, our goal was to achieve a standardized set of emotion annotations so the classification was obtained using the emotion2vec+ large classifier \cite{ma2023emotion2vec} for both VoxCeleb \cite{nagrani2020voxceleb} and Expresso \cite{nguyen2023expresso} corpora. 

\subsection{Human Annotation Refinement}\label{sec:data-human-refinement}
With partially weak transcriptions and labels for non-verbal vocalizations and emotions, we aimed to boost annotation quality through human supervision. Our objective in creating this dataset was to facilitate speech synthesis that includes non-verbal vocalizations.
To achieve high-quality data, we instructed  annotators to review and validate the annotations and to discard samples if they met any of the following conditions:

\begin{itemize}
    \item The speech is in a language other than English.
    \item There are multiple speakers talking simultaneously.
    \item Non-verbal vocalizations are produced by someone other than the main speaker (e.g., background laughter).
\end{itemize}

These filtering criteria ensured the dataset contained only high-quality samples suitable for training robust speech synthesis models that can accurately generate non-verbal elements, thus minimizing noise and improving model generalization capabilities.

Starting with the initial weak transcription, annotators were tasked with making the following edits:

\begin{itemize}
    \item Adjusting the transcription by adding, removing, or replacing words.
    \item Modifying the transcription by adding, deleting, or substituting NVs.
    \item Replacing specified emotions as needed.
\end{itemize}

Annotators were provided with detailed definitions of each NV and emotion to guide their edits. The annotation was performed on the Argilla platform \cite{Daniel_Argilla_-_Open-source_2023}.

\begin{algorithm}
\caption{Merge: combine two aligned sentences into one. }\label{alg:cap}
\begin{algorithmic}[1]
\Require $(s, t)$ - two sentences
\State $s_{align}, t_{align} \gets align(s,t)$ \algorithmiccomment{$len(s_{align}) = len(t_{align})$} 
\State $res \gets []$
\State $i \gets 0$
\While{$i < len(s_{align}$) } 
\State $c_s \gets s_{align}[i]$
\State $c_t \gets t_{align}[i]$

\If{$ c_s \neq \text{"-"} $  and $ c_t \neq  \text{"-"} $ }  \algorithmiccomment{if exists alignment}
    \State $res \gets res + [c_s]$ \algorithmiccomment{$c_s$ and $c_t$ are equal}
\ElsIf{$ c_t \neq \text{"-"} $ } 
    \State $res \gets res + [c_t]$
\ElsIf{$ c_s \neq \text{"-"} $ } 
    \State $res \gets res + [c_s]$
        \EndIf 

\State $i \gets i+1$
\EndWhile
\State \Return res
\end{algorithmic}
\end{algorithm}

\subsection{Annotation Fusion Process}
Following human annotation refinement, we obtained several transcriptions with varying annotations. In this section, we describe the process of fusing these human refinements to produce the final annotation. The core approach is to accept or decline letters or NV tags in the transcription based on majority agreement among annotators. The process consists of several steps:

\begin{enumerate}
    \item Merge all annotator hypotheses into a single comprehensive version $m_3$.
    \item Align each individual annotator's hypothesis with the merged version, receiving $a_1, a_2, a_3$.
    \item Vote on each letter and NV tag to create the final annotation $t_{target}$.
\end{enumerate}

Given the example of initial weak transcription $t_i$, the annotators' hypotheses $t_1$, $t_2$, $t_3$:
\begin{itemize}
    \item  $t_{i}$  - It's \textcolor{orange}{dog} [laugh] on the mat  
    \item   $t_{1}$ - It's a cat [laugh] on the mat  
    \item  $t_{2}$  - It's a cat [laugh] on the \textcolor{orange}{sofa}
    \item   $t_{3}$ - It's a cat \textcolor{orange}{[sigh]} on the mat    
\end{itemize}
We apply the merge algorithm~\ref{alg:cap} to create a final merged version $m_3$. The process works step by step:
\begin{equation}
    \begin{cases}
     m_{1} :=  merge(t_i, t_1) \\ 
      m_{2} :=  merge(m_{1}, t_{2}) \\
       m_{3} :=  merge(m_{2},  t_{3}) 
    \end{cases}\,.
\end{equation}
Note that the merging algorithm~\ref{alg:cap} first aligns two sequences to create two sequences of the same length. For alignment, we use the Pyalign~\footnote{https://github.com/poke1024/pyalign \label{pyalign}} library. 

Pyalign aligns text sequences by adding gap characters ("-") at strategic positions to maximize matches between sequences. It uses dynamic programming to place these gaps optimally, balancing matches, mismatches, and gap penalties. The result gives two equal-length sequences where corresponding positions show the best alignment between the original sequences, enabling effective comparison.

Table~\ref{tab:merge_example} represents the intermediate and final merge versions given the examples above.

\begin{table}[ht]
    \centering
    \resizebox{0.5\textwidth}{!}{%
    \begin{tabular}{c | c|c|c|c|c|c|c|c|c|c}
         \hline 

         $m_{1}$ &  It's  & a & dog & cat & [laugh] & - & on & the & mat & ----  \\

        $m_{2}$ &  It's  & a & dog & cat & [laugh] & - & on & the & mat & sofa  \\

         $m_{3}$ &  It's  & a & dog & cat & [laugh] & [sigh] & on & the & mat & sofa  \\
         \hline

    \end{tabular}%
    }
    \caption{The result of applying merge algorithm~\ref{alg:cap} given $t_i, t_1, t_2$ and $t_3$ as described above.}
    \label{tab:merge_example}
\end{table}

Next, we align our final merged version of the annotation $m_{3}$ with the individual annotators' hypotheses $t_1, t_2, t_3$.
\begin{equation}
    \begin{cases}
     a_{1} :=  align(m_{3}, t_1) \\ 
      a_{2} :=  align(m_{3},   t_{2}) \\
       a_{3} :=  align(m_{3},  t_{3})
    \end{cases}\,.
\end{equation}
Similarly to the merge step, we use the Pyalign\textsuperscript{\ref{pyalign}} library for alignment. 

Table~\ref{tab:annotation_example} shows the results of the alignment process. The last row indicates how many times each word or NV appears across the three alignments.
\begin{table}[ht]
    \centering
    \resizebox{0.5\textwidth}{!}{%
    \begin{tabular}{c | c|c|c|c|c|c|c|c|c|c}
        \hline
          $a_{1}$ & It's  & a & --- & cat & [laugh] & - & on & the & mat & ---- \\
        $a_{2}$ &  It's  & a & --- & cat & [laugh] & - & on & the & --- & sofa \\ 
        $a_{3}$ &   It's  & a & --- & cat & - & [sigh] & on & the & mat & ---- \\
         \hline 
         count &   3  & 3 & 0 & 3 & 2 & 1 & 3 & 3 & 2 & 1 \\
    \end{tabular}%
    }
    \caption{Alignments received from aligning $t_1, t_2, t_3$ with merged transcription $m_3$.}
    \label{tab:annotation_example}
\end{table}

The final version, $t_{target}$  is created using the majority vote algorithm~\ref{alg:mode} applied to letters and NV tags from $a_{1}, a_{2}, a_{3}$. We keep a letter (or tag) only if it appears in at least two annotators' versions, which constitutes a majority in our three-annotator setup.   

In our example, the resulting annotation is: "It's a cat [laugh] on the mat".

\begin{algorithm}[ht]
\caption{Majority vote}\label{alg:mode}
\begin{algorithmic}[1]
\Require $a_{1}, a_{2}, a_{3}$ - aligned transcriptions
\State $res \gets []$
\State $i \gets 0$
\While{$i < len(a_1$) }
\State $A = [a_1[i], a_2[i], a_3[i]]$ 
\State $c \gets mode(A)$
\If{$count(c, A) \geq 2$} 
    \State $res \gets res + [c]$
    \EndIf 
\State $i \gets i+1$
\EndWhile
\State \Return res
\end{algorithmic}
\end{algorithm}


\begin{table*}
    \centering



\begin{tabular}{lcccccccccccc}
\toprule
 & Num samples & Num speakers & Cough & Sneeze & Sigh & Breath & Laughter & Sniff & Snoring & Throat & Groan & Grunt \\
\midrule
train & 3642 & 1314 & 161 & 9 & 135 & 2760 & 890 & 186 & 11 & 104 & 111 & 7 \\
dev & 46 & 46 & 4 & 0 & 1 & 37 & 7 & 0 & 0 & 0 & 0 & 0 \\
test & 359 & 147 & 29 & 0 & 4 & 305 & 56 & 0 & 0 & 0 & 0 & 0 \\
\bottomrule
\end{tabular}

    \caption{Data statistics over train / dev / test splits. The dataset was strategically partitioned to ensure no speaker overlap between subsets, with speakers from the VoxCeleb corpus exclusively allocated for testing purposes }
    \label{tab:split_nv}
\end{table*}

\section{NVTTS Dataset}
NVTTS results from annotating and filtering existing VoxCeleb 1~\cite{Nagrani_2017}, VoxCeleb 2~\cite{Chung_2018}, and Expresso~\cite{nguyen2023expresso} corpora. This section provides an overview of the original data sources and statistics on NVs and emotion tags in the resulting dataset.

\subsection{Data Sources}
The VoxCeleb dataset was originally designed for speaker verification tasks, however, its collection of short speech clips extracted from YouTube interviews makes it an excellent potential source of nonverbal vocalizations.

Expresso is a high-quality speech dataset comprising expressively read speech and improvised dialogues. As noted in Section~\ref{sec:related-datasets}, it contains limited annotated NVs, primarily laughing and breathing, but contains a richer set of non-annotated NVs.

We selected only 0.35\% of the original VoxCeleb data, specifically segments containing NVs, while retaining 15\% of Expresso due to its naturally richer NV content. Despite this, VoxCeleb dominates the NVTTS dataset in duration (13.7 hours vs. Expresso's 3.9 hours) due to its larger overall size. Both datasets maintain a balanced gender distribution (60\% male, 40\% female). Table~\ref{tab:stats} provides detailed statistical comparisons.

\subsection{Emotion Distribution}
While our primary focus is on NVs, we also provide emotion annotations, see Section~\ref{sec:emotion_detection}. The distribution is dominated by Neutral speech, followed by Happy and an ambiguous Other category. Sadness appears in moderate amounts, while Disgust, Surprise, Anger, and Fear are relatively rare, see Table~\ref{tab:nvtts_emo}. The high prevalence of Other class suggests that our emotion taxonomy may not fully capture the nuanced variability of expressive speech. For annotation, we mainly relied on the emotion2vec+ large classifier \cite{ma2023emotion2vec}, which supports classification into 8 basic emotions. However, more recent work~\cite{diwan2025scalingrichstylepromptedtexttospeech} expands the emotional taxonomy to over 20 emotion categories, many of which may be present within our Other class. Emotion annotation remains a fundamentally challenging task. Human perception of emotion is often subjective and context-dependent, and even experienced annotators may disagree~\cite{pallewela2024optimizing}. This was also evident in our case: approximately 8.5\% of the NVTTS samples have no assigned emotion label due to a lack of agreement among annotators.

\subsection{Nonverbal Vocalization Distribution}
The NV composition mirrors real-world occurrence patterns: Breathing constitutes the majority, followed by laughter, reflecting their ubiquity in daily communication. The dataset maintains diversity across common NVs such as Sniffs, Coughs, Throat Clearing, Sighs, and Groans. Rare categories like Sneezes, Snores, and Grunts are also present, though in smaller quantities (Table~\ref{tab:nvtts_nv}). This balance ensures robust coverage of both frequent and uncommon nonverbal expressions.

\begin{table}
    \centering
\resizebox{0.47\textwidth}{!}{
\begin{tabular}{lcccccccc}
\toprule
Emo & Neutral & Happy & Other & Sad & Disgusted & Surprised & Angry & Fearful \\
\midrule
Count & 2406 & 1605 & 1177 & 387 & 78 & 34 & 25 & 10 \\
\bottomrule
\end{tabular}
}
\caption{Emotion statistics over NVTTS. We see a pronounced skew toward Neutral and Happy categories, closely mirroring the emotion distribution patterns observed in real-life speech}
\label{tab:nvtts_emo}
\end{table}

\section{Experiments}
This section presents experiments using the NVTTS dataset to train zero-shot TTS models with NV synthesis capability.

\subsection{Training and Experimental Setup}
All our experiments are based on the CosyVoice-300M~\cite{du2024cosyvoice} model.
We train only the language model component~\cite{du2024cosyvoice} of the full architecture in a supervised fine-tuning setup. We use the Adam optimizer~\cite{diederik2014adam} with a constant learning rate of $1e-5$ for 25 epochs on a single A100 GPU. The best checkpoint (after 25 epochs) was selected using automated metrics on the validation set.
Gradient accumulation is applied every 2 batches.

To validate the individual components of our dataset, we train four separate models: \textbf{(1) NVTTS-full}, trained on the full dataset; \textbf{(2) NVTTS-no-emotion}, trained without emotion labels; \textbf{(3) NVTTS-no-NV}, trained with non-verbal vocalizations removed; and \textbf{(4) NVTTS-no-emotion-no-NV}, trained with both emotion labels and non-verbal vocalizations excluded.

We benchmark against CosyVoice2~\cite{du2024cosyvoice_2}, a state-of-the-art open-source baseline capable of zero-shot generation with NVs including breath, laughter, cough, and sighs. While our base CosyVoice model initially lacked support for cough/sigh synthesis and showed inferior performance on laughter/breath generation compared to CosyVoice2, fine-tuning with NVTTS gives the opportunity to 
demonstrate NVTTS's effectiveness for enabling zero-shot TTS with comprehensive nonverbal capabilities.

\begin{figure}
\centering
\includegraphics[width=\linewidth]{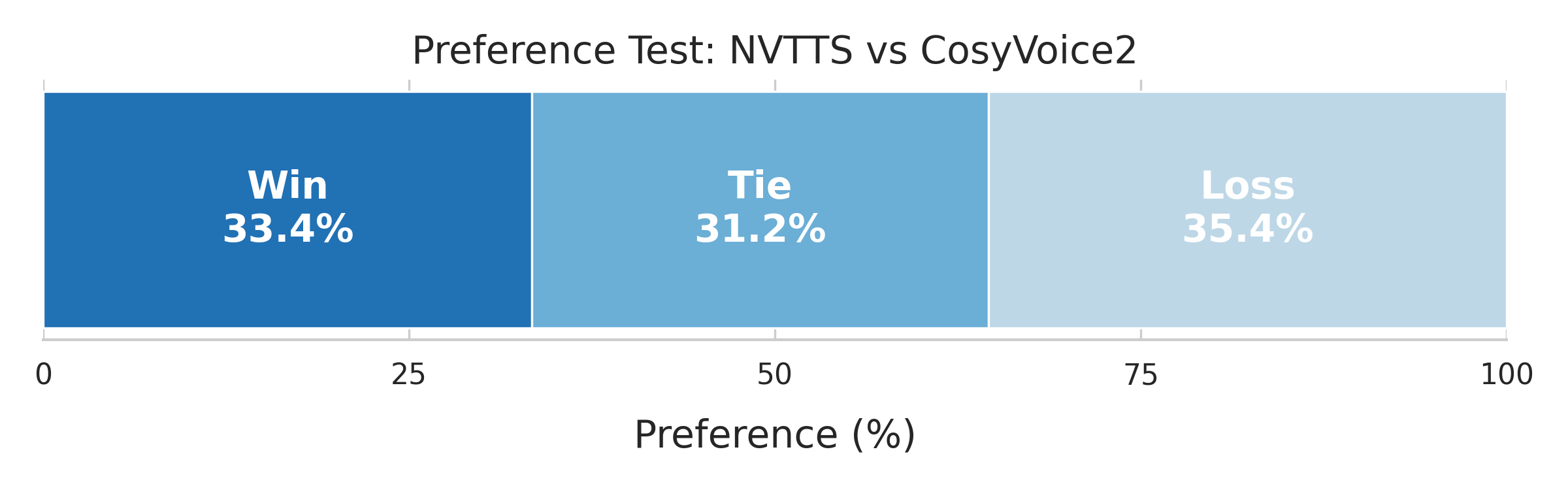}
\caption{Human evaluation results comparing the nonverbal generation capabilities of CosyVoice (trained on the NVTTS dataset) and CosyVoice2 indicate that the NVTTS-trained model achieves performance comparable to CosyVoice2, the current state-of-the-art method.
}
\label{fig}
\end{figure}

\subsection{Evaluation setup}

\begin{table}[!ht]
    \centering

    \resizebox{0.47\textwidth}{!}{
    \begin{tabular}{lcccccccc}
\toprule
 & Neutral & Happy & Other & Sad & Disgusted & Surprised & Angry & Fearful \\
\midrule
Train & 1 293 & 1 238 & 835 & 331 & 71 & 24 & 18 & 10 \\
Dev & 29 & 12 & 2 & 3 & 0 & 0 & 0 & 0 \\
Test & 220 & 68 & 71 & 6 & 0 & 0 & 1 & 0 \\
\bottomrule
\end{tabular}

}
    \caption{Emotion statistics over train / dev / test splits. The test subset contains almost only Neutral and Happy emotion tags; consequently, emotion generation capabilities were assessed mainly on the Happy category. }
    \label{tab:split_emo}
\end{table}

We evaluate model performance on both subjective and objective metrics, including word error rate (WER), speaker similarity (SIM-o)~\cite{chen2022wavlm}, emotion similarity (Emo-sim)~\cite{ma2023emotion2vec}, NV Jaccard distance~\cite{schmid2024effective}, DNSMOS~\cite{reddy2021dnsmos}, and preference test (Preference) on the test set, which includes 359 samples.

For transcribing the generated audio to compute WER, we used Whisper model~\cite{radford2022whisper}, following~\cite{du2024cosyvoice, du2024cosyvoice_2}, but specifically employed the whisper-large-v3-turbo variant. Speaker similarity was measured using wavlm-base-plus-sv \cite{chen2022wavlm} embeddings.

To compute NV Jaccard distance~\cite{schmid2024effective} we first automatically detect NVs from reference audio and generated audio using BEATs model \cite{schmid2024effective}. We then calculate the Jaccard distance between these NV sets. To evaluate performance across different vocalizations, we compute separate metrics for breathing $J_{breath}$, laugh $J_{laugh}$ and cough $J_{cough}$.

In the preference test, 3 annotators were asked to choose the better model based solely on transcriptions containing non-verbal cues (NVs), with reference audio excluded to isolate the effect of NVs.

For fair comparison with CosyVoice2~\cite{du2024cosyvoice_2} on the NVs that it can produce, we limited the NVs for testing to breathing, laughter, coughing and sighs. We randomly selected 359 test samples from Voxceleb speakers with at least two audio records, using remaining speakers' data for training. Tables~\ref{tab:split_nv} and \ref{tab:split_emo} provide detailed statistics of train, development, and test splits by NVs and emotions. 

\subsection{Results}

\subsubsection{Automatic Evaluation Results}

Table~\ref{tab:automatic_res} presents the automatic evaluation results. Our models outperform in speaker similarity (SIM-o) and emotion similarity (EMO-SIM), likely due to NVTTS dataset's greater speaker and emotion diversity. While intelligibility remains comparable between models, CosyVoice2 achieves higher DNSMOS scores, suggesting better sound quality—possibly reflecting noisier samples in our training data.

Both systems perform equally well in cough generation. NVTTS shows slight advantages in synthesizing breathing, while CosyVoice2 excels at laughter generation. This discrepancy likely stems from tokenization differences: CosyVoice2 distinguishes between single-token laughter ([laugh]) and multi-token spans ($<$laughter$></$laughter$>$), whereas our implementation lacks this granularity.

\subsubsection{Ablation Study: Impact of Tags}

Our ablation study on emotion and NVs tags usage revealed two key findings: First, removing emotion annotations marginally improves NV generation quality according to automatic metrics, leading us to select NVTTS (w/o emo) for further comparisons. Second, eliminating NV tags degrades NV detection accuracy for both settings (with and without emotion). This demonstrates the essential role of explicit NV modeling in our experimental setup. See Table~\ref{tab:automatic_res} for more details.

\subsubsection{Human Evaluation}
To ensure fair comparison, we evaluated CosyVoice2 (w/o emo) against NVTTS (w/o emo) in a human side-by-side study (Figure~\ref{fig}). Annotators were instructed to prioritize audio-transcript alignment (including NVs) while disregarding speaker/emotion similarity. 

The original CosyVoice2 model was preferred slightly more often (35.4\% vs 33.4\%). However, the Wilson score interval (95\% CI, continuity-corrected \cite{2020SciPy-NMeth}) for CosyVoice2's success rate is [30.4, 40.6], indicating no statistically significant difference between  our NVTTS-trained  model, and the original CosyVoice2 model (p $>$ 0.05), which was trained on a proprietary in-house dataset.

\begin{table}[ht]
    \centering
    \resizebox{0.47\textwidth}{!}{%
\begin{tabular}{lcccccccc}
\toprule
 & SIM-o \textuparrow & WER \textdownarrow & EMO-SIM \tablefootnote{We calculate EMO-SIM on emotional subset of test (i.e without neutral and other)} \textuparrow  & DNSMOS \textuparrow & $J_{cough}$ \textuparrow & $J_{breath}$ \textuparrow & $J_{laugh}$ \textuparrow & $J$ \textuparrow \\
Model &  &  &  &  &  &  &  &  \\
\midrule

cosyvoice2-full & 0.85 & 0.22 & 0.52 & 3.9 & 0.17 & 0.74 & 0.15 & 0.63 \\

cosyvoice2-no-emotion & 0.75 & \textbf{0.18} & 0.50 & \textbf{3.93} & 0.19 & 0.89 & \textbf{0.34} & 0.78 \\

\toprule

NVTTS-full  & \textbf{0.89} & 0.21 & 0.56 & 3.82 & 0.17 & 0.91 & 0.23 & 0.79 \\

NVTTS-no-emotion& \textbf{0.89} & 0.19 & 0.57 & 3.82 & \textbf{0.2} & \textbf{0.92} & 0.25 & \textbf{0.8} \\

NVTTS-no-NV  & \textbf{0.89} & 0.22 & \textbf{0.58} & 3.82 & 0.17 & 0.89 & 0.19 & 0.76 \\

NVTTS-no-emotion-no-NV & \textbf{0.89} & \textbf{0.18} & \textbf{0.58} & 3.82 & 0.15 & 0.88 & 0.2 & 0.76 \\

\bottomrule
\end{tabular}
}
 
    \caption{We compute automatic evaluation metrics on model inference results in a zero-shot setting using the test split of NVTTS dataset.}
    \label{tab:automatic_res}
\end{table}

\section{Conclusion}

This work addresses a critical bottleneck in expressive speech synthesis -- the lack of high-quality, open-source datasets with diverse NVs. To this end, we introduce NVTTS, a 17-hour corpus enriched with 10 NV types and 8 emotion categories, providing a solid foundation for training and benchmarking prosodically expressive TTS systems. Our reproducible annotation pipeline — combining automated NV detection, emotion classification, and human validation — ensures precise alignment with transcriptions, while maintaining diversity in speakers and emotional content. 

Despite its modest size compared to modern large-scale corpora, NVTTS offers a valuable starting point for developing scalable NV and emotion annotation techniques. Future work may leverage it to investigate whether speech encoders implicitly capture nonverbal cues and to develop retrieval systems that identify NV-rich segments from larger unannotated datasets. In this way, NVTTS can facilitate the training and validation of automated annotators, serving as a seed for semi-supervised discovery of nonverbal data at scale. 

By open-sourcing both the dataset and the annotation methodology, we aim to alleviate a longstanding barrier in the field and support reproducible research in expressive, context-aware speech generation. Dataset available at  \url{https://huggingface.co/datasets/deepvk/NonverbalTTS}


\bibliographystyle{IEEEtran}
\bibliography{mybib}

\end{document}